%% file: ISA.tex
\documentclass[review]{elsarticle}

\usepackage{lineno,hyperref}
\modulolinenumbers[50]

\journal{Pattern Recognition}

\usepackage{xspace}

\newcommand{\IntegrateM}{Integrated Sequence Autoencoder\xspace}
\usepackage{multirow}
\usepackage{subfigure} 
\usepackage{amsmath}
\usepackage{amssymb}
\usepackage{comment}
\usepackage[table]{xcolor}
\usepackage{booktabs}
\definecolor{gray}{gray}{0.8}

\usepackage{lipsum}
\makeatletter
\def\ps@pprintTitle{%
	\let\@oddhead\@empty
	\let\@evenhead\@empty
	\def\@oddfoot{}%
	\let\@evenfoot\@oddfoot}
\makeatother

\begin{document}

\begin{frontmatter}

\title{Unsupervised Learning of Sequence Representations by Autoencoders}

\author[mainadd]{Wenjie Pei\corref{mycorrespondingauthor}}
\ead{wenjiecoder@gmail.com}
\cortext[mycorrespondingauthor]{Corresponding author}

\author[mainadd]{David M.J. Tax}
\ead{D.M.J.Tax@tudelft.nl}
\address[mainadd]{Pattern Recognition Laboratory, Delft University of Technology}


%

\begin{abstract}
	\input{Abstract.tex}
\end{abstract}


\end{frontmatter}


\section{Introduction}
\input{Introduction.tex}

\section{Related Work}
\input{related_work.tex}

\section{Model}
\input{Model.tex}

\section{Experiments}
\input{Experiments.tex}

\section{Conclusion}
\input{Conclusion.tex}

\section*{References}

\bibliography{ref}

\end{document}

%% file: Abstract.tex
Sequence data is challenging for machine learning approaches, because the lengths of the sequences may vary between samples.
In this paper, we present an unsupervised learning model for sequence data, called the Integrated Sequence Autoencoder (ISA), to learn a fixed-length vectorial representation by minimizing the reconstruction error. Specifically, we propose to integrate two classical mechanisms for sequence reconstruction which takes into account both the global silhouette information and the local temporal dependencies. Furthermore, we propose a stop feature that serves as a temporal stamp to guide the reconstruction process, which results in a higher-quality representation. The learned representation is able to effectively summarize not only the apparent features, but also the underlying and high-level style information. Take for example a speech sequence sample: our ISA model can not only recognize the spoken text (apparent feature), but can also discriminate the speaker who utters the audio (more high-level style). 
One promising application of the ISA model is that it can be readily used in the semi-supervised learning scenario, in which a large amount of unlabeled data is leveraged to extract high-quality sequence representations and thus to improve the performance of the subsequent supervised learning tasks on limited labeled data.

%% file: Introduction.tex
Sequence data is ordered in space or time, which is common around us such as video, audio or text data. Traditional machine learning models for fixed length observations are challenged by sequence data. An important reason is that the length of sequence examples is variable. Typically, sequence models are specifically designed to deal with sequence data for extensive applications ranging from computer vision~\cite{Karpathy2014,TAGM} to natural language processing~\cite{Bahdanau14}. However, 
it is not an efficient way to redesign the entire existing single-observation models to cater to sequence data. It would be  beneficial if we can transform sequence data to the standard fixed-length feature representation, which can then be readily accessed by any existing traditional machine learning models. 

In this paper, we aim to learn a fixed-length representation given a sequence. To be applicable to both labeled and unlabeled sequence data, the representation should be learned in an unsupervised way. Besides, the obtained representation should be able to capture the informative features contained in the sequence. 
One straightforward way 
is to employ the Dynamic Time Warping (DTW) algorithm~\cite{DTW_vintsyuk,DTW_sakoe}. To be specific, we can first construct a sequence vocabulary for reference and then calculate the pairwise distances from a target sequence sample to all the samples in the vocabulary. The achieved distance vector can be used as the representation to describe this target sample.  Since DTW is good at aligning two sequences which share a similar shape but may vary in speed, this representation can capture the shape information well. However, it is difficult for DTW to explore more latent features that is not related to the global shape. For instance, while DTW can easily identify the audio content, it is difficult for DTW to recognize the speaker who utters the audio.

Here we introduce an unsupervised learning model for sequences, called Integrated Sequence Autoencoder (ISA), to learn a fixed-length vectorial representation. Specifically, our model integrates seamlessly two classical mechanisms for sequence reconstruction that takes into account both the global and local features. The representation is learned by minimizing the reconstruction error and the whole model can be trained in an end-to-end manner efficiently. Extensive experiments demonstrated that the learned representation by our model can not only capture the apparent shape information as well as DTW, but also summarize the underlying and high-level style information that DTW does not capture.

One key difference that distinguishes our model from the classical sequence-to-sequence supervised models used for machine translation~\cite{Bahdanau14} or text generation~\cite{LiLJ15} is that those models predict words out of a vocabulary as a classification task in decoding. An ending token is added into the vocabulary to indicate a full stop during the sequence generation (reconstruction). In contrast, our model reconstructs the input sequence as a regression task. We predict/reconstruct a continuous feature value for each time step. It is hard for our model to perceive the temporal progress during decoding. To address this issue, we propose a stop feature. This acts as a temporal stamp to guide the reconstruction process. It can potentially help the model to better align the reconstructed sequence with the input one, which improves the quality of the representation. 
\vspace{-5pt}



%% file: related_work.tex
It is straightforward to employ Dynamic Time Warping (DTW) algorithm~\cite{DTW_vintsyuk,DTW_sakoe} to extract representation for a sequence by calculating the pairwise distance to a vocabulary of sequences. While the representation obtained in this way is able to capture the global shape information well due to the specialty of DTW,  it is challenging for it to explore more latent features which is not related to shape information. Here we propose to combine the ideas from autoencoders and sequence-to-sequence learning. We discuss related work on both topics separately below.

\smallskip\noindent \textbf{Autoencoder} is a type of neural network that is designed to learn a latent representation  for samples by minimizing the reconstruction error~\cite{autoencoder}. 
It can be used for dimension reduction~\cite{HinSal06} or unsupervised pre-training~\cite{Erhan2010}. Various variations of autoencoders have been proposed to improve the quality of learned representation, such as Denoising  autoencoder~\cite{SDA}, Sparse autoencoder~\cite{SparseAE} or Variational Autoencoder~\cite{VAE}.  Autoencoder is typically used for single-observation tasks, which means the length of input is fixed. In contrast, our model focuses on learning representation for sequence data whose length is variable.

\smallskip\noindent\textbf{Sequence-to-sequence Learning} aims to learn a mapping function from an input sequence to an output sequence. It has been extensively studied in many tasks like machine translation~\cite{Bahdanau14,cho-emnlp14} or text generation~\cite{LiLJ15,Graves2013GeneratingSW}. It is generally achieved by first encoding the input sequence to a latent representation and then decoding it into an output sequence~\cite{Sutskever2014}. Typically, the decoding phase is formulated as a supervised classification task, making it hard to predict real-valued sequences. Our ISA model tries to avoid this and uses a regression formulation.
\vspace{-6pt}

%% file: Model.tex
\begin{figure}[tb]
	\includegraphics[width=1\linewidth]{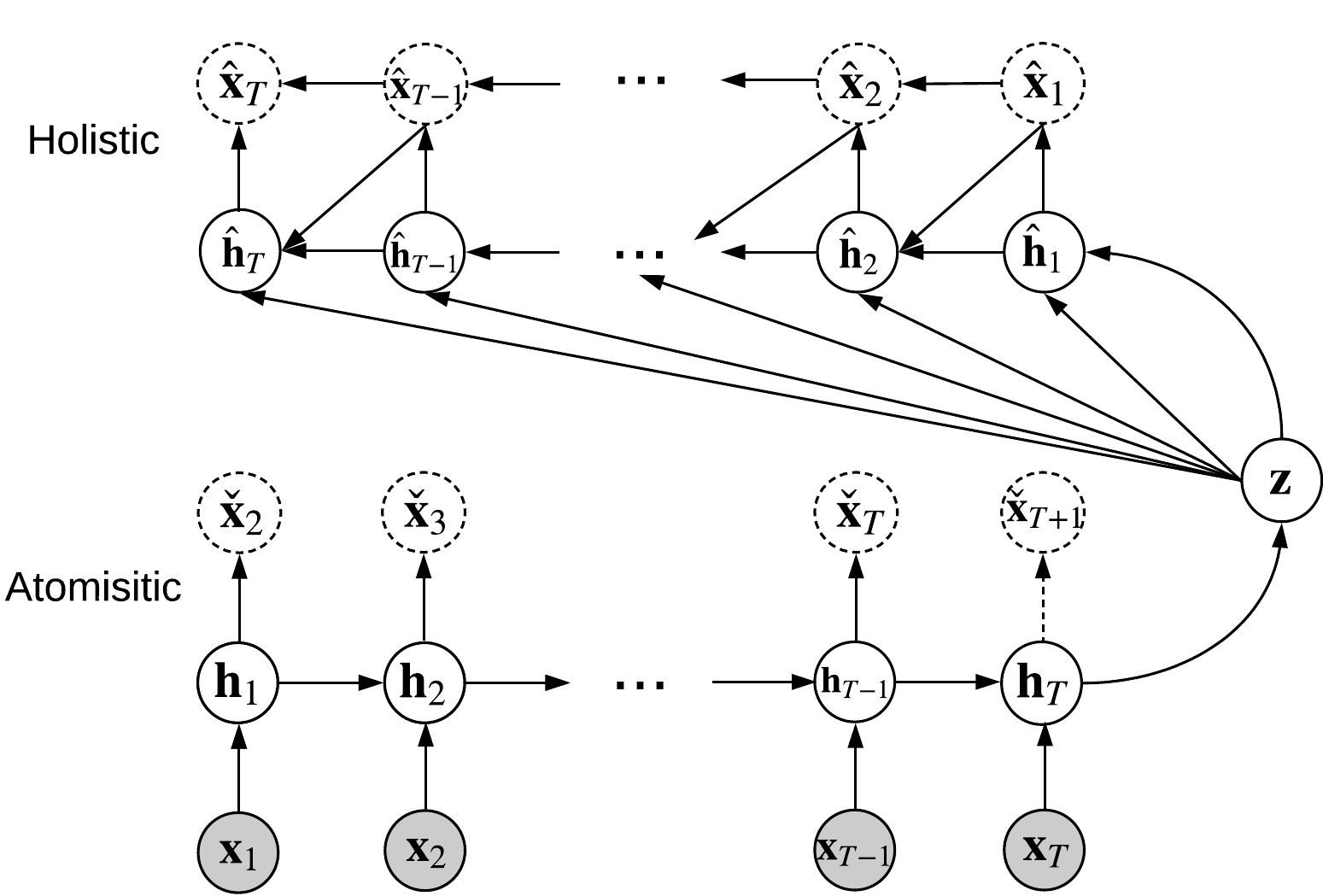}
	\centering
	\caption{The graphical structure of our \IntegrateM. Note that $\check{\mathbf{x}}_{T+1}$ is a virtual node not used in practice.}
	\label{fig:architecture}
\end{figure}

Given an input sequence of variable length, our goal is to learn a vectorial representation in an unsupervised way. The learned representation  is required to: (1) have fixed length, which can be readily used for further processing tasks, (2) summarize high-level features covering not only the holistic information contained in the whole sequence, but also the temporal dependencies between observations. Our proposed model is composed of two core modules: a Holistic autoencoder and an Atomistic autoencoder. They are integrated seamlessly via a loss function and the whole model can be trained in an end-to-end manner efficiently. We will elaborate on these two modules first and then present our \IntegrateM. The graphical structure of the model is illustrated in Figure~\ref{fig:architecture}. 

\subsection{Holistic Sequence Autoencoder}
The goal of Holistic Sequence Autoencoder is to learn a hidden representation which is able to capture the holistic features for the given sequence. This is achieved by reconstructing the full sequence from only one hidden representation vector $\mathbf{z}$.
Similar to the classical autoencoder, The Holistic Sequence autoencoder consists of two parts: (1) the encoder which is used to encode the input data into a hidden representation $\mathbf{z}$, and (2) the decoder for reconstructing the input sequence from the hidden representation. We employ Long Short-Term Memory (LSTM)~\cite{LSTM} as the backbone network for both the encoder and decoder due to its powerful modeling for the sequence data.

\paragraph{\textbf{Encoder}}
The encoder of the Holistic Sequence Autoencoder is in line with the routine forward pass of LSTM model. Formally, given an input sequence $\mathbf{x}_{1, \dots, T} = \{\mathbf{x}_1, \dots, \mathbf{x}_T\}$ of length $T$ in which $\mathbf{x}_t \in\mathbb{R}^D$ denotes the observation at the $t$-th time step of the sequence, the hidden state $\mathbf{h}_t$ at time step $t$ of the encoder is modeled as the element-wise product $\odot$ between the output gate $\mathbf{o}_t$  of the LSTM at time step $t$ and the cell state $\mathbf{C}_t$ (transformed by $\tanh$):
\begin{equation}
\mathbf{h}_{t} = \mathbf{o}_t \odot \tanh(\mathbf{C}_t),
\label{eqn:encoder_h}
\end{equation} 
where the cell state $\mathbf{C}_t$  is calculated by a weighted sum of the candidate cell state at the current time step $\mathbf{\tilde{C}}_t$ and the cell state in the previous step $\mathbf{C}_{t-1}$:
\begin{equation}
\mathbf{C}_{t} = \mathbf{f}_t \odot \mathbf{C}_{t-1} + \mathbf{i}_t \odot \mathbf{\tilde{C}}_t.
\label{eqn:encoder_C}
\end{equation} 
Herein, $\mathbf{f}_t$ and $\mathbf{i}_t$ are respectively the forget gate and input gate to control the information flow from the previous and current  time step. All three gates ($\mathbf{i}_t$, $\mathbf{f}_t$, $\mathbf{o}_t$) and the candidate cell state $\mathbf{C}'_t$ are modeled based on the previous hidden state and current input observation in a similar nonlinear way. For instance,  the output gate $\mathbf{o}_t$ (at $t$-th time step) is computed by:
\begin{equation}
\mathbf{o}_t = \sigma (\mathbf{W_o} \cdot \mathbf{h}_{t-1} + \mathbf{U_o} \cdot \mathbf{x}_t +b_o).
\label{eqn:encoder_s}
\end{equation} 
Herein, $\mathbf{W}_o$ and $\mathbf{U}_o$ are the transformation matrices and $b_o$ is the bias term. The hidden state $\mathbf{h}_T$ at the last time step is employed as the summary representation $\mathbf{z}$ for the sequence that we aim for: 
\begin{equation}
\mathbf{z} = \mathbf{h}_T.
\end{equation}

\paragraph{\textbf{Decoder}}
To optimize the learning of the hidden representation $\mathbf{z}$, our Holistic Sequence Autoencoder attempts to reconstruct the whole input sequence $\mathbf{x}_{1, \dots, T}$ from $\mathbf{z}$ and minimize the reconstruction error to make the reconstructed sequence as close as possible to the original sequence. As a result, the learned representation $\mathbf{z}$ is encouraged to preserve all the informative features contained in the sequence. Specifically, we employ another LSTM model as our decoder to reconstruct the sequence step by step. The reconstructed observation $\hat{\mathbf{x}}_t$ at time step $t$ is modeled as:
\begin{equation}
\hat{\mathbf{x}}_{t} = \mathbf{M} \cdot g (\mathbf{A} \cdot \mathbf{z} + \mathbf{B} \cdot \hat{\mathbf{x}}_{t-1} + \mathbf{E} \cdot \hat{\mathbf{h}}_t),
\end{equation} 
which takes into account the summary representation $\mathbf{z}$, the hidden state $\hat{\mathbf{h}}_t$ in current step and the reconstructed observation in the previous step $\hat{\mathbf{x}}_{t-1}$.  Here $\mathbf{M}, \mathbf{A}$, $\mathbf{B}$ and $\mathbf{E}$ are mapping matrices and a rectified linear unit (ReLU) function~\cite{ReLU} is used as the activation function $g$. The hidden state $\hat{\mathbf{h}}_t$ is derived similarly to the hidden state $\mathbf{h}_t$ of the Encoder (Equation~\ref{eqn:encoder_h} and \ref{eqn:encoder_C}). It should be noted that the calculation of the three gates ($\hat{\mathbf{i}}_i$, $\hat{\mathbf{f}}_t$, $\hat{\mathbf{o}}_t$) and candidate cell state  $\hat{\mathbf{C}'}_t$ is slightly different from Equation~\ref{eqn:encoder_s}, since they also rely on the summary representation $\mathbf{z}$ apart from the hidden state $\hat{\mathbf{h}}_t$ and reconstructed observation $\hat{\mathbf{x}}_{t-1}$. For any $\hat{\mathbf{s}} \in \{\hat{\mathbf{i}}_i, \hat{\mathbf{f}}_t, \hat{\mathbf{o}}_t, \hat{\mathbf{C}}'_t\}$, it is modeled as:
\begin{equation}
\hat{\mathbf{s}} = \sigma (\hat{\mathbf{W}} \cdot \hat{\mathbf{h}}_{t-1} + \hat{\mathbf{U}} \cdot \hat{\mathbf{x}}_{t-1} + \hat{\mathbf{V}} \cdot \mathbf{z} + \hat{b}).
\label{eqn:decoder_s}
\end{equation} 

\paragraph{\textbf{Parameter Learning}} Suppose we are given a training set $\mathcal{D} = \{ \mathbf{x}^{(n)}_{1, \dots, T^{(n)}}\}_{n=1,\dots, N}$ containing $N$ sequences of length $T^{(n)}$. We learn all parameters involved in a Holistic Sequence Autoencoder jointly by minimizing the reconstruction error. In particular, we minimize the mean squared error between the reconstructed sequence and the original input sequence for each training sample in $\mathcal{D}$:
\begin{equation}
\mathbf{L}_h = \sum_{n=1}^{N} \left( \frac{1}{T^{(n)}} \sum_{t=1}^{T^{(n)}}\lVert \hat{\mathbf{x}}^{(n)}_t - \mathbf{x}^{(n)}_t \rVert^2 \right).
\label{eqn:holistic_loss}
\end{equation} 
\subsection{Atomistic Sequence Autoencoder}
In contrast to the Holistic Sequence Autoencoder, which specializes in capturing the holistic features, the Atomistic Sequence Autoencoder is designed to focus on modeling the local temporal dependencies between adjacent steps in a sequence. It also employs a LSTM model as its backbone, but the difference from Holistic Sequence Autoencoder is that it performs encoding and decoding using only one LSTM module. As illustrated in Figure~\ref{fig:architecture}, we first encode the input observation $\mathbf{x}_t$ at time step $t$ into the hidden state $\mathbf{h}_t$, which is exactly same as the Holistic Sequence Autoencoder (Equation~\ref{eqn:encoder_h}, \ref{eqn:encoder_C} and \ref{eqn:encoder_s}). Then the hidden state $\mathbf{h}_t$ is decoded to reconstruct the observation $\check{\mathbf{x}}_{t+1}$ for the next time step:
\begin{equation}
\check{\mathbf{x}}_{t+1} = \mathbf{P} \cdot g (\mathbf{F} \cdot \mathbf{h}_t),
\end{equation}  
where $\mathbf{P}$ and $\mathbf{F}$ are linearly mapping matrices.
Similar to the Holistic Sequence Autoencoder, all the parameters of Atomistic Sequence Autoencoder are learned by minimizing the reconstruction error for the training samples:
\begin{equation}
\mathbf{L}_a = \sum_{n=2}^{N} \left( \frac{1}{T^{(n)}} \sum_{t=1}^{T^{(n)}}\lVert \check{\mathbf{x}}^{(n)}_t - \mathbf{x}^{(n)}_t \rVert^2 \right).
\label{eqn:atomistic_loss}
\end{equation} 
Since the observation at each time step is reconstructed from the hidden state in the previous time step, the Atomistic Sequence Autoencoder tends to focus on the accuracy of local transitions between time steps in the reconstructed sequence rather than the global outline.
\subsection{Integrated Sequence Autoencoder}
To assimilate the merits of both Holistic Sequence Autoencoder and Atomistic Sequence Autoencoder, we propose to integrate the two autoencoders. The resulting model is called Integrated Sequence Autoencoder (ISA). Figure~\ref{fig:architecture} shows that two autoencoders can be seamlessly integrated by sharing one LSTM module. The shared LSTM module is not only used for performing encoding of both two models, but also responsible for the decoding of the Atomistic Sequence Autoencoder. The Integrated Sequence Autoencoder is trained in an end-to-end manner by computing the weighted sum of Loss functions of two autoencoders:
\begin{equation}
\mathbf{L} = \alpha \cdot \mathbf{L}_h + (1-\alpha) \cdot \mathbf{L}_a,
\label{eqn:integrated_loss}
\end{equation} 
where $\alpha \in [0, 1]$ is a hyper-parameter that balances the impact of two sub-autoencoders.
\subsection{Stop Feature}

\begin{figure}[htb]
	\includegraphics[width=0.65\linewidth]{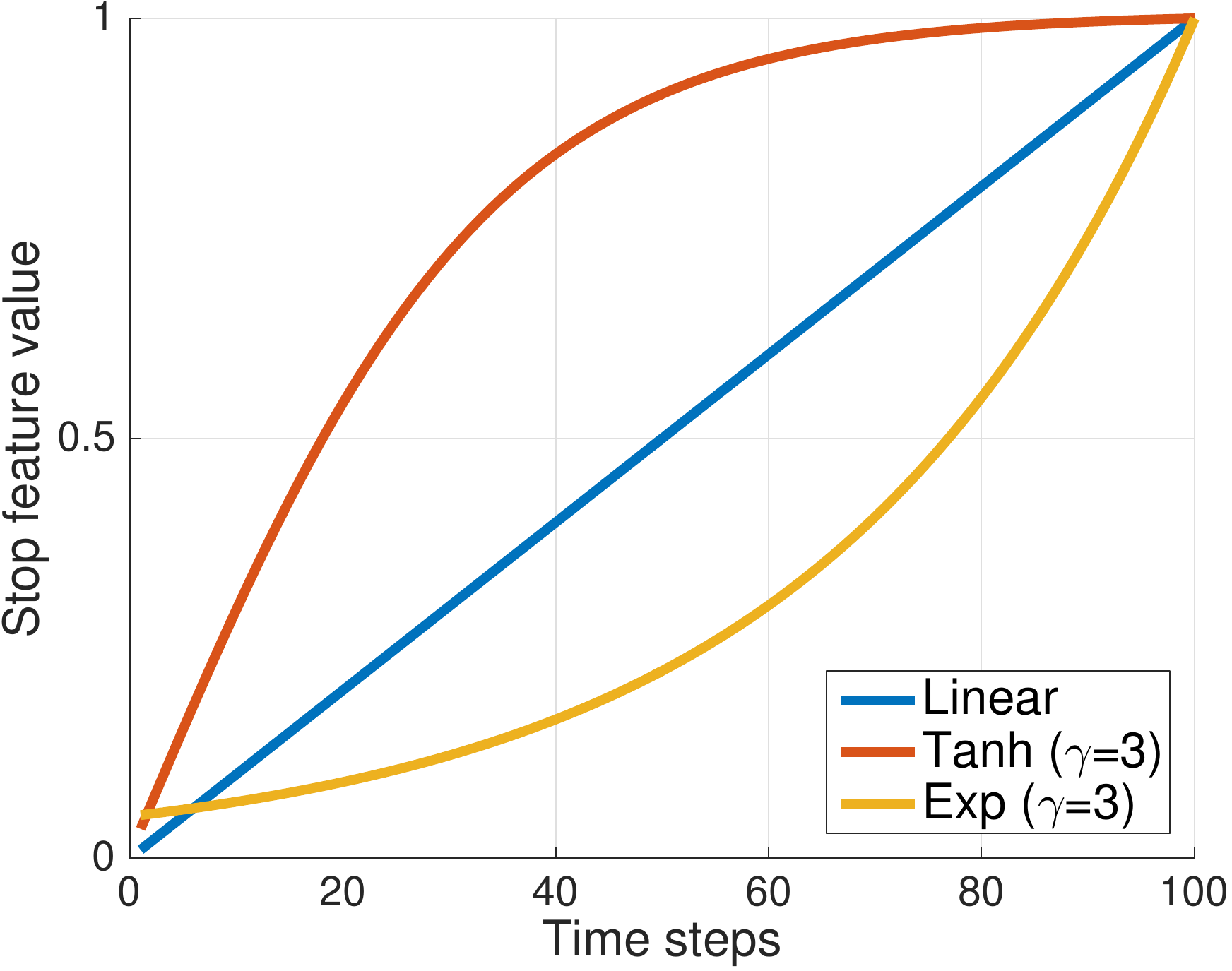}
	\centering
	\caption{Three mechanisms for stop feature. }
	\label{fig:stop_feature}
\end{figure}

To help our model track the temporal progress during decoding, we propose a stop feature to guide the reconstruction process. Specifically, we concatenate a scalar value $v_t\in [0,1]$ to the feature vector for each time step $[\mathbf{x}_t v_t]^T$. The value of the stop feature increases smoothly as the time step moves forward. Higher value (closer to $1$) indicates a closer position to the end of the sequence. Hence the stop feature acts like a temporal stamp stick to each time step of a sequence. We investigate three different mechanisms for the stop feature (shown in Figure~\ref{fig:stop_feature}):
\begin{itemize}
\item \textbf{Linear} mechanism: the stop feature value increases linearly with time. Mathematically, the stop feature $v_t$ at time step $t$ of a sequence with total length $T$ is defined as:
\begin{equation}
v_t = \frac{t}{T}.
\end{equation}
\item \textbf{Tanh} mechanism: the increase of $v$ is rapid at the beginning of the sequence, but slows down at the end. The stop feature $v_t$ at time step $t$ is modeled as:
\begin{equation}
v_t = \tanh \left(\gamma \cdot \frac{t}{T}\right) + 1 - \tanh(\gamma),
\label{eqn:tanh}
\end{equation}
where $\gamma > 0$ is a hyper-parameter to determine the gradient. 
\item \textbf{Exp} mechanism: the stop feature value increases exponentially. The stop feature $v_t$ at time step $t$ is modeled as (for $\gamma>0$):
\begin{equation}
v_t = \exp \left(\gamma \cdot \frac{t-T}{T}\right).
\label{eqn:exp}
\end{equation}
\end{itemize}
Intuitively, the stop feature can potentially help the model to memorize the relative temporal location and thus better align the reconstructed sequence with the input sequence. Consequently, it leads to higher quality of learned representation, which is demonstrated by experiments in Section~\ref{sec:stop_feature}.

%% file: Experiments.tex
\label{sec:experiment}
We first conduct experiments to compare our ISA model to the DTW algorithm, in which we aim to investigate what is learned by the representations of both models. Then we perform comparisons between the Holistic, Atomistic models and our proposed ISA model, and validate our theoretical analysis regarding their respective advantages. Subsequently, experiments are performed to evaluate the effectiveness of the stop feature and  its effect on sequence encoding. Then, we evaluate our model by feeding the learned representation to a  classifier (i.e., a Support Vector Machine, SVM) and compare the classification performance to other state-of-the art sequence classification models. Finally, we conduct experiments to show that our model can be potentially applied to semi-supervised learning scenario, in which the unlabeled data is leveraged by our model to enhance the learned representation and thus to improve the performance of classification tasks on limited labeled data. 
\subsection{Experimental Setup}
\subsubsection{Datasets}
We conduct experiments on three datasets, selected to show generalization across different tasks and modalities: (1) a online handwritten character dataset (OHC)~\cite{OHC}, (2) a audio dataset of Arabic spoken digits~\cite{ASD} and (3) the Cohn–Kanade extended facial expression dataset (CK+)~\cite{CK}.

The online handwritten character (OHC) data set~\cite{OHC} is a pen-trajectory sequence dataset that comprises three dimensions of features at each time step: the pen movement in the x and y directions, and the pen pressure. The dataset contains 2858 sequences with an average length of 120 time steps. Each sequence corresponds to a handwritten character that has one of 20 labels. 

The Arabic spoken digit dataset~\cite{ASD} consists of 8800 utterances, which were collected by asking 88 Arabic speakers to utter each of 10 digits ten times.  Each sequence comprises 13-dimensional Mel-Frequency Cepstral Coefficients (MFCCs), which were sampled at 11,025Hz, 16 b using a Hamming window. We use two different versions of this dataset: (1) a \emph{digit} version in which the uttered digit is the class label (10 classes) and (2) a \emph{voice} version in which the speaker is the class label (88 classes). 

The Cohn–Kanade extended facial expression dataset (CK+)~\cite{CK} consists of 593 image sequences (videos) from 123 subjects.
Each video presents a single facial expression. We use a subset of 327 videos in our experiments, which have validated labels corresponding to one of seven emotions (anger, contempt, disgust, fear, happiness, sadness, and surprise).  We adopt the shape features used in~\cite{maaten2012action} as the feature representation. Each frame is represented by the variation of 68 feature point locations
(x, y) with respect to the first frame~\cite{CK}, which results in 136-dimensional feature representation for each frame in the video.
\subsubsection{Settings}
We perform 5-fold cross validation on OHC dataset, and 10-fold cross validation on CK+ dataset across all the experiments. For the Arabic (digit) dataset, we use the same data division as Hammami and Bedda~\cite{ASD}: 6600 samples as training set and 2200 samples as test set. There is no speaker overlap between them. Regarding the Arabic (voice), we use the samples of the first 8 digits as training set and the left out samples as test set. 

For all the recurrent networks mentioned in this work, the number of hidden units is tuned by selecting the best configuration from the candidate set $\{32, 64, 128\}$ with a validation set set side from the training set. The loss function weight $\alpha$ of \IntegrateM (Equation~\ref{eqn:integrated_loss}) and the hyper-parameter $\gamma$ of the Stop feature in Equation~\ref{eqn:tanh} and~\ref{eqn:exp} are tuned by minimizing the reconstruction error on the validation set. We perform gradient descent optimization using RMSprop~\cite{RMSprop} to train our model. The gradients are clipped in the interval $[-5, 5]$~\cite{gradientclip}. As a way to evaluate our model and DTW, we perform classification on the learned representations using SVM classifier~\cite{svm,libsvm} with radial basis kernel. The kernel parameter and regularization parameter of the SVM are tuned with the validation set. 

\subsection{Comparison to DTW}
\vspace{-2pt}
The most intriguing question to investigate is whether or not our model learns the same information as DTW. 
We perform experiments on the Arabic (digit) and Arabic (voice) datasets. Specifically, given a test sample, we employ DTW to calculate the pairwise distances from it to all training samples as its 
DTW-based representation, whose dimensionality is therefore the number of training samples. Then we perform comparisons between this DTW-based representation and the representation learned by our model, both in supervised and unsupervised ways.  

\begin{figure}[htb]
	\includegraphics[width=0.75\linewidth]{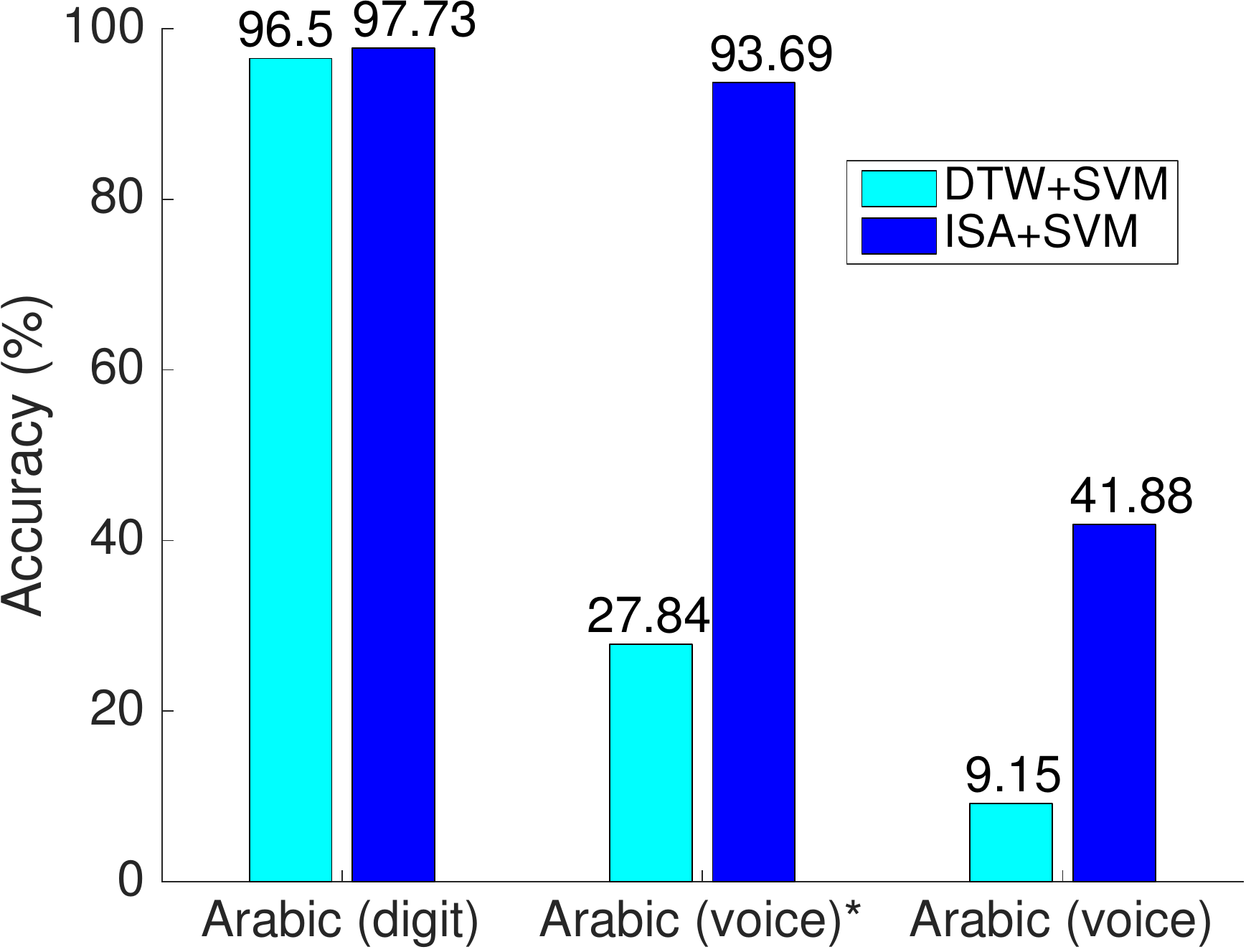}
	\centering
	\caption{The comparison of classification performance of SVM on representations learned by DTW and \IntegrateM (ISA) with stop feature on Arabic (digit) and Arabic (voice) datasets.  Note that Arabic (voice) has no digit overlap between training and test set while all 10 digits can be seen in both training and test sets for Asterisked Arabic voice* dataset.}
	\label{fig:ISA_DTW_bar}
\end{figure}

\paragraph{\textbf{Evaluation of Classification performance}} Typically, a better representation results in better classification performance. We first evaluate the classification performance of both representations (from DTW and our model) by feeding them to the same classifier, i.e., an SVM with the same settings. The experimental results are presented in Figure~\ref{fig:ISA_DTW_bar}. 
Both models perform quite well on Arabic (digit) dataset, which is a relatively easy task. It is not surprising for DTW since it is good at aligning two sequences which share a similar shape but may vary in speed, which fit well with digit classification. However, it fails to capture the discriminative information for voice (speaker) classification while our ISA model significantly outperforms DTW. It demonstrates that our model is able to learn the high-level representation which captures not only the global shape feature but also the underlying characteristics. Herein, classification on Arabic (voice) is much more challenging than Arabic (voice)* since there is no digit overlap between training and test sets for Arabic (voice).

\begin{figure}[htb]
	\begin{center}
		$\begin{array}{cc}
		\includegraphics[width=0.47\linewidth]{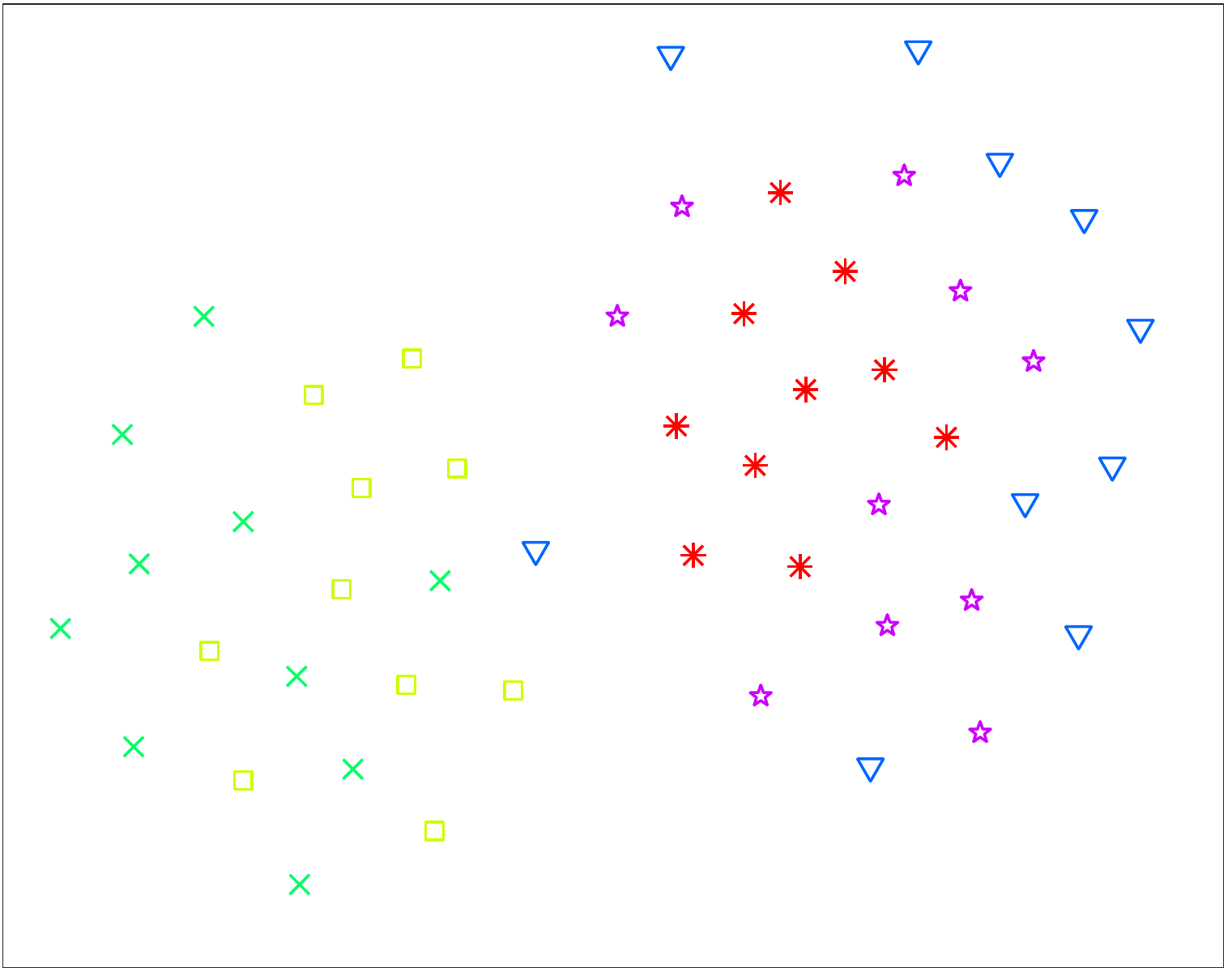} &
		\includegraphics[width=0.47\linewidth]{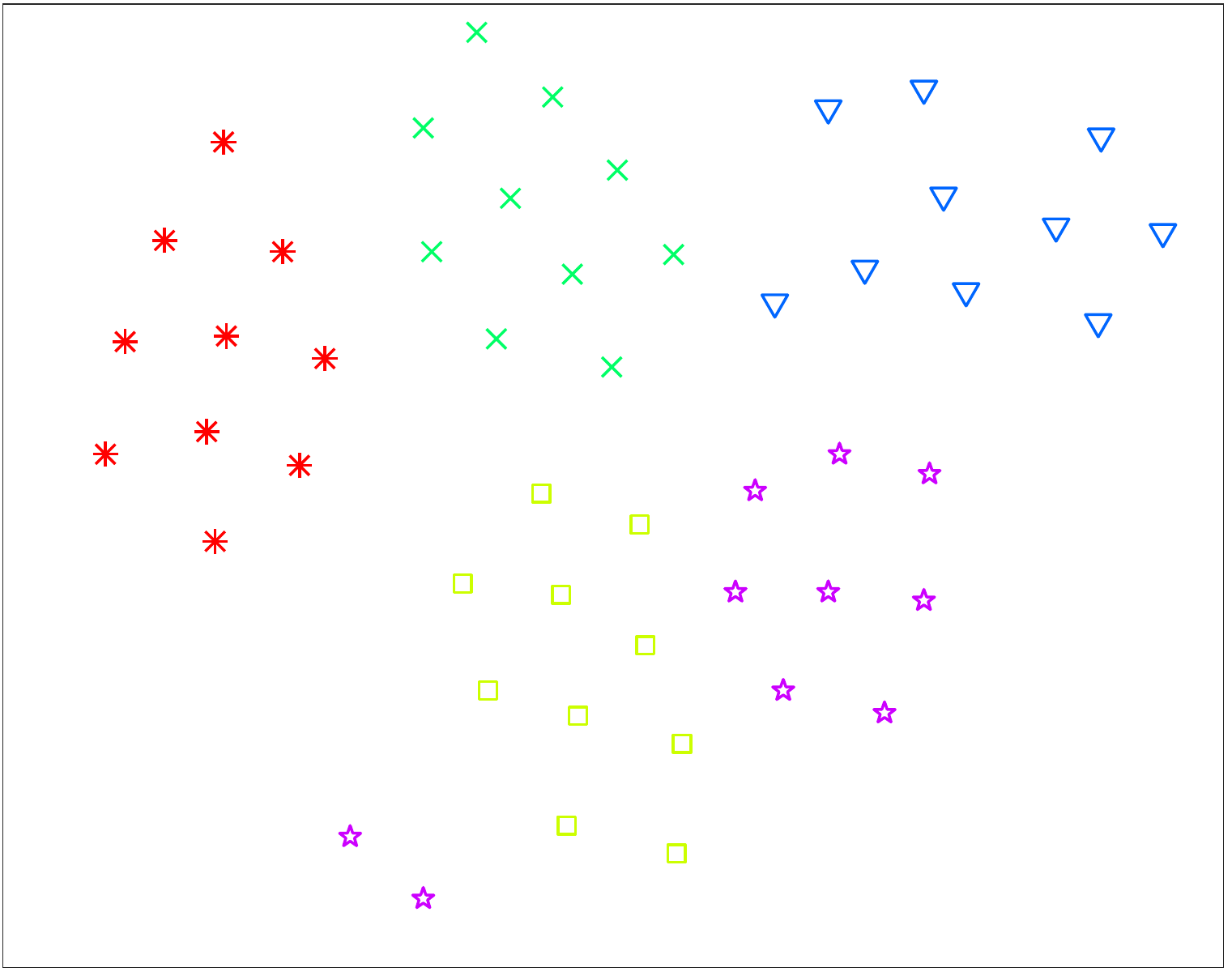} \\
		\text{(a) DTW} & \text{(b) ISA}
		\end{array}$
	\end{center}
	\caption{t-SNE maps of the Arabic (voice) test data for the same digit uttered by 5 randomly selected speakers (indicated by different colors and markers), constructed based on (a) DTW distance (to each training sample) and (b) our \IntegrateM (ISA). The DTW distance measure fails to discriminate the samples between voice of different speakers while our \IntegrateM is able to handle it well.}
	\label{fig:tsne_comparison}
\end{figure}

\paragraph{\textbf{Clustering Visualization}} To obtain more insight into what representation both models have been learned on the Arabic (voice) dataset, we apply t-SNE~\cite{tsne}. Figure~\ref{fig:tsne_comparison} shows t-SNE maps of 50 utterances of the same digit from 5 randomly selected speakers for both models. In the DTW two main clusters can be distinguished, but within a cluster speakers are confused. The ISA representation, on the other hand more strongly focus the individual speakers (although not perfectly: the cluster of the speaker indicated by `$\star$' misses two samples). 

\subsection{Comparison of Different Autoencoders}
Next we compare the representations learned by the Holistic, Atomistic and Integrated sequence autoencoders. In particular, we perform classification on the learned representations with an SVM on four different tasks. The experimental results in Table~\ref{table:autoencoder_comparison} show that the Holistic autoencoder outperforms the Atomistic on the Arabic (digit) dataset whilst Atomistic exhibits better performance on voice classification. It is consistent with the theoretical analysis that Holistic Autoencoder specializes in capturing the global feature while the Atomistic autoencoder focuses on the local temporal information. Similarly, Holistic autoencoder performs much better than Atomistic autoencoder on OHC dataset whose discriminative feature is mainly concerning the global shape of a sequence. The ISA, which is designed to assimilate the merits of both Holistic and Atomistic autoencoders, performs best among all the tasks.

\begin{table}[!htb]
	\caption{Classification accuracy (\%) on 4 datasets by performing SVM on the learned representation of different autoencoders. The best performance per dataset is shown in bold.}
	\begin{center}
		\renewcommand\arraystretch{1.2}
		\begin{tabular}{l|cc|c}
			\toprule
			\multirow{2}{*}{\textbf{Dataset}} & \multicolumn{3}{c}{\textbf{Model}}
			\\&  \textbf{Atomistic} & \textbf{Holistic} & \textbf{ISA}  \\ 
			\midrule
			\rowcolor{gray}Arabic (digit) &68.36 & 96.23& \textbf{97.41}    \\
			Arabic (voice)& 32.10 & 26.88& \textbf{35.63}\\
			\rowcolor{gray} OHC& 53.01 & 92.26 &\textbf{94.15}\\
            CK+  & 96.43 & 92.50& \textbf{96.79} \\
			\bottomrule
		\end{tabular}
	\end{center}
	\label{table:autoencoder_comparison}
\end{table}

\begin{figure}[htb]
	\begin{center}
		$\begin{array}{c|c}
		\includegraphics[width=0.47\linewidth]{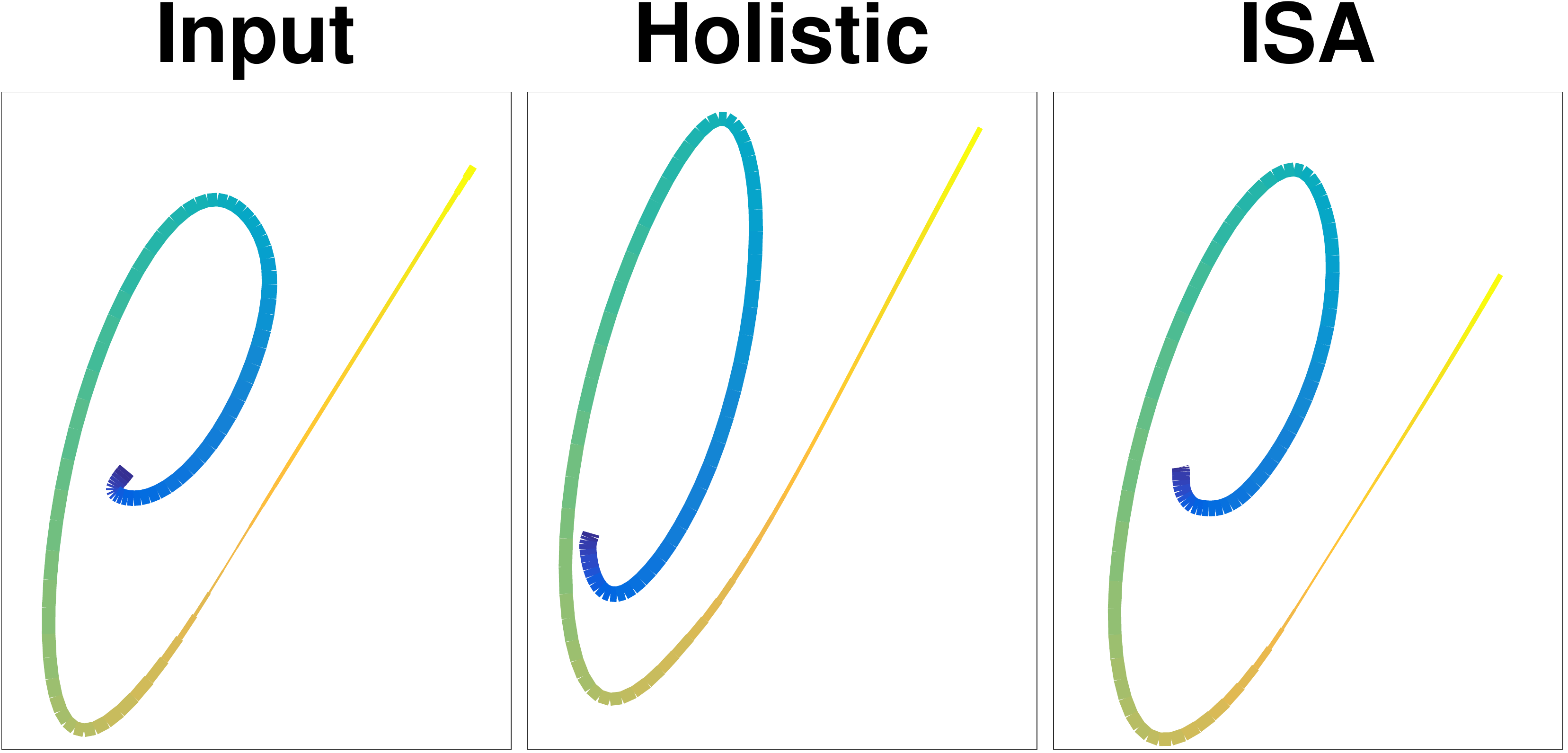} &
		\includegraphics[width=0.47\linewidth]{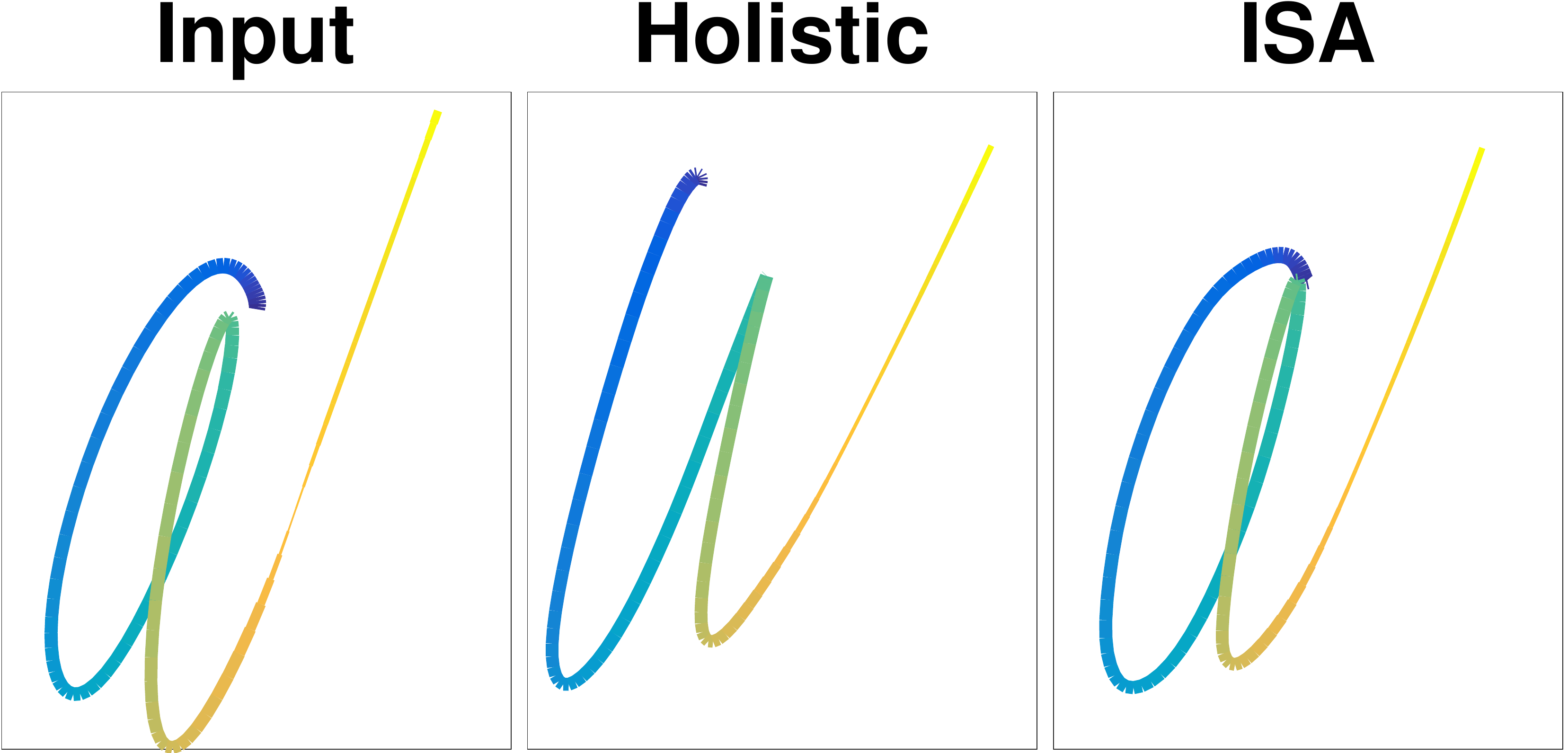} \\
		\text{(1) character `e'} & \text{(2) character `a'}
		\end{array}$
	\end{center}
	\caption{Visualization of some (hard) reconstructed characters (in OHC dataset) by Holistic autoencoder and Integrated Sequence Autoencoder (ISA). Both subfigures shows that ISA is able to reconstruct more accurate sequences than Holistic autoencoder. The pen trajectory moves from blue color to yellow color. Thicker line indicates higher pen pressure.}
	\label{fig:autoencoder_vis}
\end{figure}

To qualitatively evaluate the performance of reconstruction, we visualize the reconstructed sequence of character examples from OHC dataset. Figure~\ref{fig:autoencoder_vis} presents two examples which are not well handled by Holistic autoencoder, probably because it is confused between similar handwritten `e' and `o', also between handwritten 'a' and `u'. It shows that the ISA is able to reconstruct more accurate sequences than Holistic autoencoder.

\subsection{The Importance of the Stop Feature}
\label{sec:stop_feature}
\paragraph{\textbf{Validation of functionality with synthetic dataset}}  
We first validate the functionality of the stop feature with a synthetic dataset. We construct the synthetic data by drawing circles with the same radius but with a different number of loops (repetitions). Specifically, two classes of circle samples are constructed: the first class consists of 100 circle samples with 2 loops and the other class contains 100 samples with 3 loops. Each sample is represented as a sequence of 2-dimensional features ($x$ and $y$ coordinates) for each time step. The variable length of sequences is sampled from a uniform distribution within an interval $[50, 200]$.

We learn the representations of the circle dataset using our \IntegrateM, with and without stop features respectively. The representations are fed into the SVM classifier to perform classification. Table~\ref{table:stop_feat} shows that ISA without any stop feature performs poorly on discriminating between circles of 2 loops and circles of 3 loops. It is not surprising since the number of loops is the only discriminative information available, and this is hardly captured by the baseline ISA (without stop feature). In contrast, all three different stop features enable ISA to classify the circles of different loops perfectly.

\begin{table}[!htb]
	\caption{Classification accuracy (\%) on 4 datasets by performing SVM on the learned representation of \IntegrateM (ISA) with and without the stop features. The best performance per dataset is boldfaced. }
	\begin{center}
		\renewcommand\arraystretch{1.2}
				\resizebox{0.8\linewidth}{!}{
		\begin{tabular}{l|c|ccc} \toprule
			\multirow{2}{*}{\textbf{Dataset}} & \multicolumn{4}{c}{\textbf{Model}}
			\\&  \textbf{ISA} & \textbf{ISA+Linear}&\textbf{ISA+Tanh} & \textbf{ISA+Exp}  \\ 
			\midrule
			\rowcolor{gray}Circle (synthetic) &77.5 & \textbf{100}& \textbf{100} &  \textbf{100}   \\
			\midrule
			Arabic (digit) &97.41 & 97.05& \textbf{97.73}& 96.86   \\
			\rowcolor{gray}Arabic (voice)& 35.63 & \textbf{41.88}&36.48 &37.61\\
			 OHC& 94.15 & 93.12 &\textbf{95.01} &94.66\\
			\bottomrule
		\end{tabular}
	}
	\end{center}
	\label{table:stop_feat}
\end{table}

\paragraph{\textbf{Classification performance on real datasets}}
Next we conduct similar classification experiments on 3 real datasets to evaluate the stop feature. The experimental results in table~\ref{table:stop_feat} show that overall the stop feature \emph{Tanh} and \emph{Linear} perform better than \emph{Exp}.
\begin{figure}[htb]
	\begin{center}
		\includegraphics[width=0.7\linewidth]{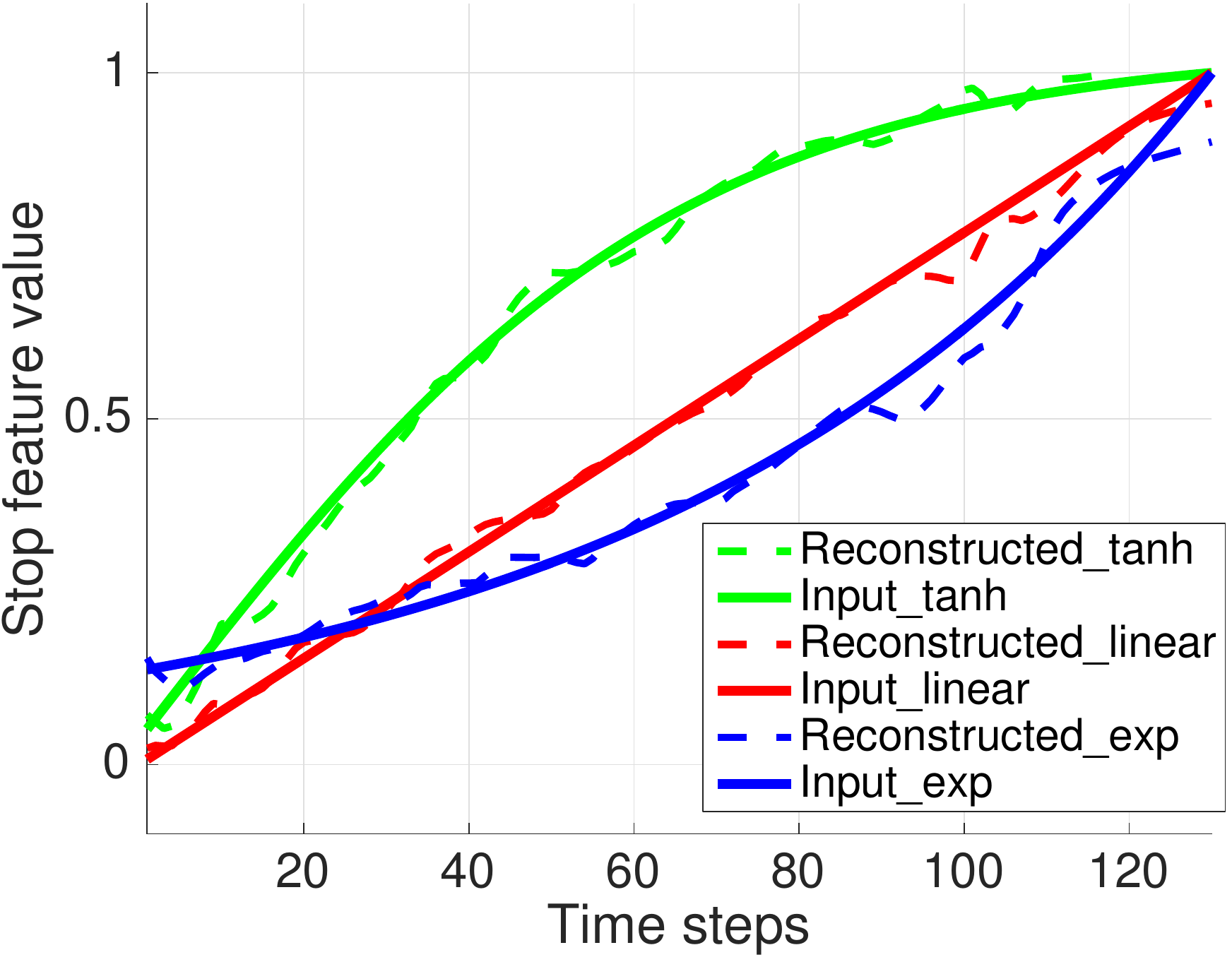} 
	\end{center}
	\caption{Visualization of the stop feature of a reconstructed character sample (in OHC dataset) by Integrated Sequence Autoencoder (ISA). The dashed lines indicate the reconstructed values and solid lines are the ground truth (input stop feature values).}
	\label{fig:stop_feat_vis}
\end{figure}

To gain more insight, we visualize the stop feature values of a reconstructed character sample by  the \IntegrateM with three mechanisms in Figure~\ref{fig:stop_feat_vis}. It shows that the reconstructed stop features are generally aligned well with the ground truth. Furthermore, the \emph{Tanh} mechanism performs more stable and accurate than the other two, especially at the tail of the sequence. It is probably because the \emph{Exp} grows exponentially, hence it is getting harder for the autoencoder to predict the values 
at tail part of the sequence.  It is consistent with the quantitative results in Table~\ref{table:stop_feat} that the stop feature \emph{Tanh} and \emph{Linear} overall outperform \emph{Exp}.

\subsection{Comparison to State-of-the-art Sequence Classifiers}
In this set of experiments, we evaluate our model by feeding the learned representation to an SVM classifier  and compare the classification performance to other state-of-the art sequence classifiers. It is actually not completely fair because these sequence classifiers train their model using label information while our model has to learn the representation in an unsupervised way. 

The sequence classifiers we use as baseline models are: (1) Hidden-Unit Logistic Model (HULM)~\cite{HULM}, which is a graphical model for sequence classification employing binary hidden units to model latent structure in the data, and (2) LSTM-based classifier, which is mounted a softmax layer and cross entropy loss upon the hidden representation of last time step of LSTM to perform classification. Furthermore, the DTW-based representation (with the SVM classifier) is also taken into comparison.

Table~\ref{table:classification} presents the classification performance of 4 classifiers on 5 different tasks. Our model and DTW+SVM perform very well on Arabic (digit) task and even outperform the other two dedicated sequence classifiers. It indicates that both our model and DTW can capture the shape information of the digits properly. In contrast, DTW performs fairly poor on the Arabic (voice) task while our model achieves a comparable performance with LSTM model. It is quite a challenging task since the training and test set have no digit overlap, and the models have to generalize to new unseen digits. Our model (as well as DTW) has to learn and preserve both discriminative digit information and voice information simultaneously in one representation 
while the supervised sequence classifiers are taught by the labels to focus on the voice classification due to the steering of the label information.

We then investigate an easier case (Arabic (voice*)) in which all 10 digits appear in both the training and test set. In this scenario, 
the performance of our model is on par with the  state-of-the-art (by HULM). Regarding the CK+ dataset whose discriminative information about facial expression is concerning both global and local features, our model achieves the best result and substantially outperforms other models while DTW does not perform well. It demonstrates the advantage of our model over DTW that our model is able to capture the features in both holistic and atomistic levels.
\begin{table}[htb]
	\caption{Classification accuracy (\%) on all 5 problems by different models. Note that there is no digit overlap between training and test set in Arabic (voice) data while all 10 digits appear in both training and test sets for Asterisked Arabic voice* dataset.}
	\begin{center}
		\renewcommand\arraystretch{1.2}
				\resizebox{1.0\linewidth}{!}{
		\begin{tabular}{l|ccc|c}
			\toprule
			\multirow{2}{*}{\textbf{Dataset}} & \multicolumn{4}{c}{\textbf{Model}}
			\\&   \textbf{HULM} & \textbf{LSTM} & \textbf{DTW+SVM}  & \textbf{ISA-Stop + SVM} \\ 
			\midrule
			\rowcolor{gray}Arabic (digit) &95.32  & 96.05 & 96.50&   \textbf{97.73} \\
			Arabic (voice)  &  \textbf{56.14} &  43.75& 9.15& 41.88 \\
			\rowcolor{gray}Arabic (voice)*   & \textbf{94.55} &  93.01 & 27.84&   93.69\\
			OHC & 97.66 ($\pm1.15$) & \textbf{98.08} ($\pm1.48$) &97.88 $(\pm{1.08}$)& 95.01 ($\pm0.65$) \\
			\rowcolor{gray}CK+  &  93.56 ($\pm4.83$)&  94.64 ($\pm3.47$)& 80.36 ($\pm6.13$)& \textbf{96.43 ($\pm4.12$)}\\
			\bottomrule
		\end{tabular}
	}
	\end{center}
	\label{table:classification}
\end{table}

\subsection{Leveraging the unlabeled data for semi-supervised learning}
\begin{figure}[htb]
	\includegraphics[width=0.85\linewidth]{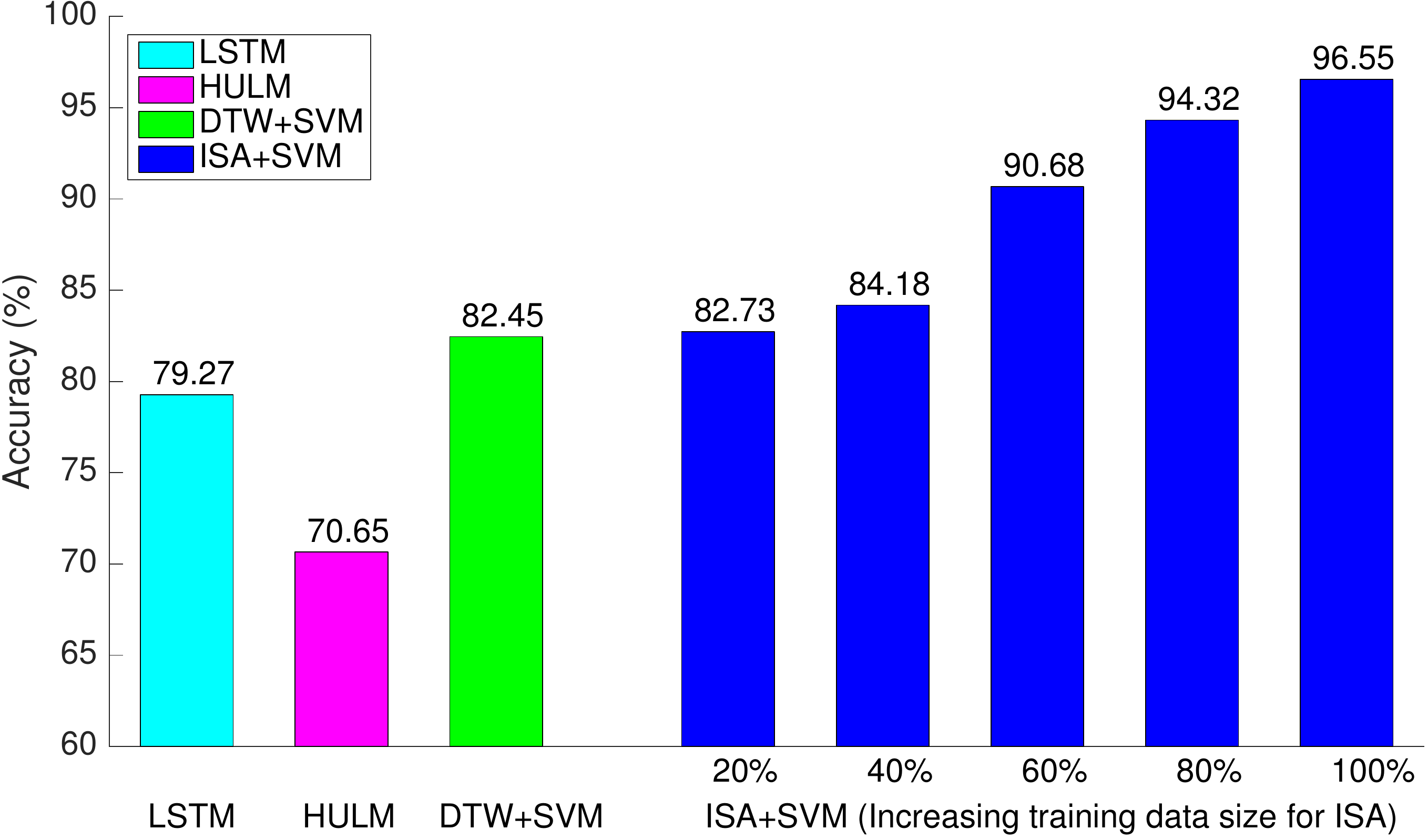}
	\centering
	\caption{The Simulation of leveraging the unlabeled data for semi-supervised learning scenario on Arabic (digit) dataset with our ISA model. All classifiers (LSTM, HULM, SVM) are trained with 20\% training data. To simulate the semi-supervised learning scenario, our ISA model is trained with increasing training data size (20\% to 100\%) without using labels while the associated SVM is trained with the same subset of 20\% (labeled) training data constantly. }
	\label{fig:semisupervised_bar}
\end{figure}

Since our model works in unsupervised way, it can be potentially used in a semi-supervised learning scenario. Specifically, we can employ our model to learn high-quality sequence representations by training it on a large amount of unlabeled data which is easily obtained, and then train a classifier (like SVM) on an available labeled dataset which is typically small. Figure~\ref{fig:semisupervised_bar} presents a simulation of such semi-supervised learning scenario for our model. We consider 20\% of the training data as the labeled data and the left 80\% are unlabeled. We train our ISA model with increasing training data (without using labels!) while training the SVM classifier with the same subset of labeled 20\% data constantly. We also list the classification performance for other classifiers (LSTM, HULM and DTW+SVM) as a reference. 

We observe that increasing the (unlabeled) training data for our model can substantially improve the classification performance of the subsequent SVM, which indicates that leveraging the unlabeled data indeed improves the quality of learned representations by our model. This is actually a fairly promising result, since our model provides a feasible way of leveraging a large amount of unlabeled sequence data for feature refinement to improve the performance of supervised learning tasks, which cannot be directly achieved by typical supervised learning methods.

%% file: Conclusion.tex
In this work, we present the Integrated Sequence Autoencoder (ISA), an unsupervised learning model to learn a fixed-length representation for varying-length sequences. The model combines the ideas from autoencoders and sequence-to-sequence learning, and integrates two classical mechanisms of sequence reconstruction. The representation learned by our model is able to capture both the global shape information and local temporal dependencies contained in the sequence data. Furthermore, we propose a stop feature to guide the reconstruction process, which helps the model to better align the reconstructed sequence with the input sequence and thus improves the quality of the learned representation. We show the generalization of our approach on three datasets that are across different tasks and modalities. As future work, we intend to extend our model in such a way that the model is able to adaptively determine when to stop during decoding  rather than reconstructing a sequence of the same length with the input sequence. Besides, 
the denoising of the sequence by reconstruction could be a promising direction.